\documentclass[10pt,twocolumn,letterpaper]{article}

\usepackage{iccv}
\usepackage{times}
\usepackage{epsfig}
\usepackage{graphicx}
\usepackage{amsmath}
\usepackage{amssymb}

\usepackage{bm}
\usepackage{booktabs}
\usepackage{siunitx}
\usepackage{stfloats}
\usepackage{multirow}
\usepackage{textcomp} 
\usepackage{pifont}
\usepackage[utf8]{inputenc}

\newcommand{\xmark}{\ding{86}}

\usepackage[pagebackref=true,breaklinks=true,letterpaper=true,colorlinks,bookmarks=false]{hyperref}

\iccvfinalcopy 

\def\iccvPaperID{3217} 

\ificcvfinal\fi

\begin{document}

\title{MicroNet: Improving Image Recognition with Extremely Low FLOPs}
\author{Yunsheng Li\textsuperscript{1}, Yinpeng Chen\textsuperscript{2}, Xiyang Dai\textsuperscript{2}, Dongdong Chen\textsuperscript{2}, Mengchen Liu\textsuperscript{2},\\
Lu Yuan\textsuperscript{2}, Zicheng Liu\textsuperscript{2}, Lei Zhang\textsuperscript{2}, Nuno Vasconcelos\textsuperscript{1} \\
\textsuperscript{1} University of California San Diego
\textsuperscript{2} Microsoft \\
{\tt\small\texttt{\{yul554,nvasconcelos\}@ucsd.edu,}}\\
{\tt\small\texttt{\{yiche,xidai,dochen,mengcliu,luyuan,zliu,leizhang\}@microsoft.com}}}

\maketitle
\ificcvfinal\thispagestyle{empty}\fi
\begin{abstract}
   This paper aims at addressing the problem of substantial performance degradation at extremely low computational cost (e.g. 5M FLOPs on ImageNet classification). We found that two factors, sparse connectivity and dynamic activation function, are effective to improve the accuracy. The former avoids the significant reduction of network width, while the latter mitigates the detriment of reduction in network depth. 
   Technically, we propose micro-factorized convolution, which factorizes a convolution matrix into low rank matrices, to integrate sparse connectivity into convolution. We also present a new dynamic activation function, named Dynamic Shift Max, to improve the non-linearity via maxing out multiple dynamic fusions between an input feature map and its circular channel shift. 
   Building upon these two new operators, we arrive at a family of networks, named MicroNet, that achieves significant performance gains over the state of the art in the low FLOP regime. For instance, under the constraint of 12M FLOPs, MicroNet achieves 59.4\% top-1 accuracy on ImageNet classification, outperforming MobileNetV3 by 9.6\%. Source code is at \href{https://github.com/liyunsheng13/micronet}{https://github.com/liyunsheng13/micronet}.
   
\end{abstract}

\section{Introduction}
Recent progress in efficient CNN architectures \cite{squeezenet16,howard2017mobilenets,sandler2018mobilenetv2,Howard_2019_ICCV_mbnetv3,Zhang_2018_CVPR,ma_2018_ECCV,tan-ICML19-efficientnet} successfully decreases the computational cost of ImageNet classification from 3.8G FLOPs (ResNet-50 \cite{he2016deep}) by two orders of magnitude to about 40M FLOPs (e.g. MobileNet, ShuffleNet), with a reasonable performance drop. However, they suffer from a significant performance degradation when reducing computational cost further. For example, the top-1 accuracy of MobileNetV3 degrades substantially from 65.4\% to 58.0\% and 49.8\% when the computational cost drops from 44M to 21M and 12M MAdds, respectively. In this paper, \textit{we aim at improving accuracy at the extremely low FLOP regime from 21M to 4M MAdds}, which marks the computational cost decrease of another order of magnitude (from 40M).

\begin{figure}[t]
	\begin{center}
		\includegraphics[width=0.75\linewidth]{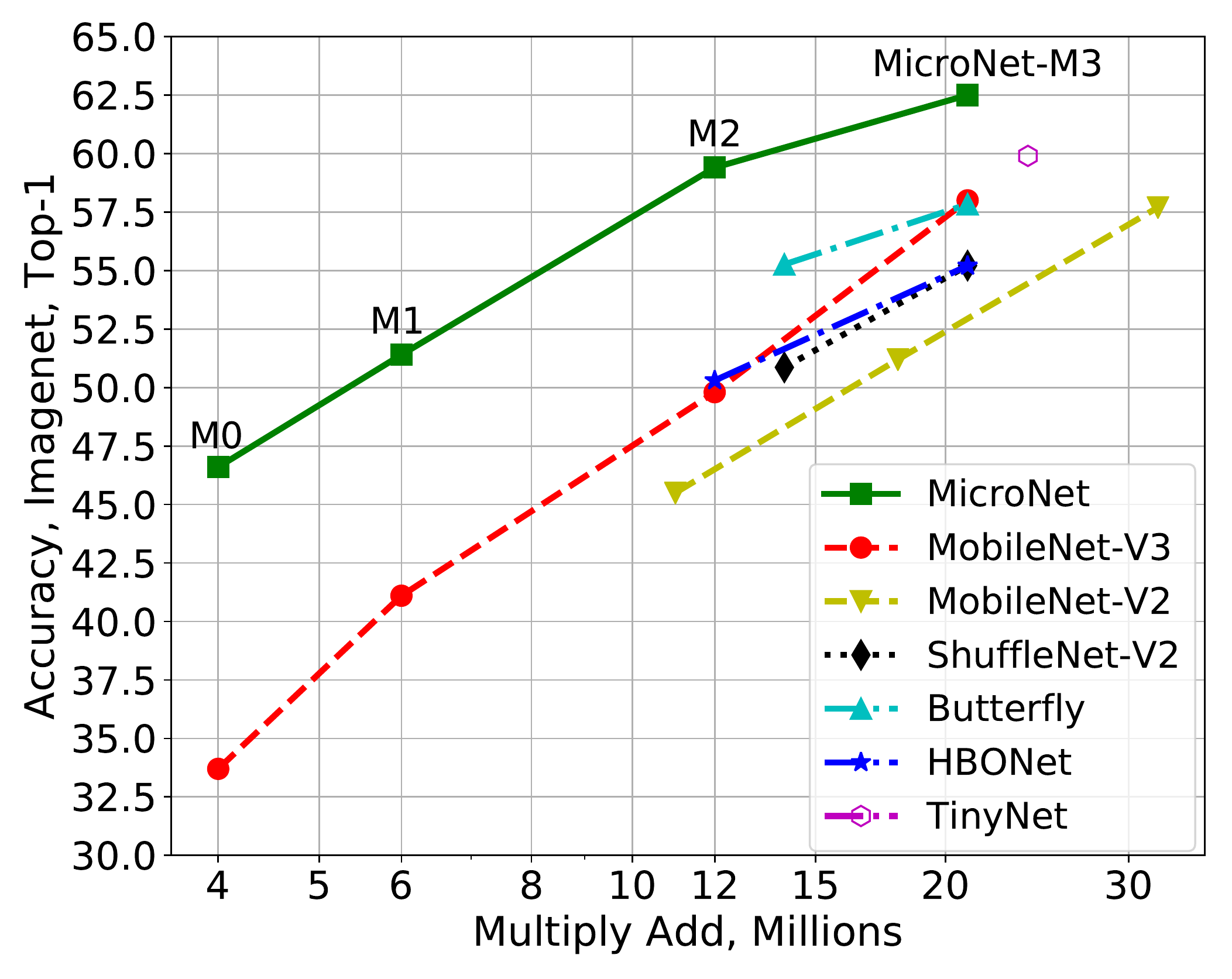}
	\end{center}
	\vspace{-3mm}
	\caption{\textbf{Computational Cost (MAdds) vs. ImageNet Accuracy.} MicroNet significantly outperforms the state-of-the-art efficient networks at very low FLOPs (from 4M to 21M MAdds).}
	\label{fig:micronet-teasor}
	\vspace{-3mm}
\end{figure}

The problem of dealing with extremely low computational cost (4M--21M FLOPs) is very challenging, considering that 2.7M MAdds are consumed by a thin stem layer that contains a single $3 \times 3$ convolution with 3 input channels and 8 output channels over a $112 \times 112$ grid (stride=2). The remaining resources are too limited to design the convolution layers and 1,000 class classifier required for effective classification. As shown in Figure \ref{fig:micronet-teasor}, a common strategy to reduce the width or depth of existing efficient CNNs (e.g. MobileNet \cite{howard2017mobilenets, sandler2018mobilenetv2, Howard_2019_ICCV_mbnetv3} and ShuffleNet \cite{Zhang_2018_CVPR, ma_2018_ECCV}) results in a severe performance degradation. Note that we focus on new operator design while fixing the input resolution to 224$\times$224 even for the budget of 4M FLOPs.

In this paper, we handle the extremely low FLOPs from two perspectives: \textit{node connectivity} and \textit{non-linearity}, which are related to the network width and depth. First, we show that lowering node connectivity to enlarge network width provides a good trade-off for a given computational budget. Second, we rely on improved layer non-linearities to compensate for reduced network depth, which determines the non-linearity of the network. These two factors motivate the design of more efficient convolution and activation functions.

Regarding convolutions, we propose a \textit{Micro-Factorized convolution (MF-Conv)} to factorize a pointwise convolution into two group convolution layers, where the group number $G$ adapts to the number of channels $C$ as:  
\begin{align}
G = \sqrt{C/R}, \nonumber
\end{align}
where $R$ is the channel reduction ratio in between. As analyzed in Section \ref{micro-fac-point-conv}, this equation achieves a good trade-off between the number of channels and node connectivity for a given computational cost. 
Mathematically, the pointwise convolution matrix is approximated by a block matrix ($G \times G$ blocks), whose blocks have rank-1. This guarantees minimal path redundancy (with only one path between any input-output pair) and maximum input coverage (per output channel), enabling more channels implementable by the network for a given computational budget.

With regards to non-linearities, we propose a new activation function, named \textit{Dynamic Shift-Max (DY-Shift-Max)}, which non-linearly fuses channels with dynamic coefficients. In particular, the new activation forces the network to learn to fuse different circular channel shifts of the input feature maps, using coefficients that adapt to the input, and to select the best among these fusions. This is shown to enhance the representation power of the group factorization with little computational cost. 

Based upon the two new operators (MF-Conv and DY-Shift-Max), we obtain a family of models, called \textit{MicroNets}. Figure \ref{fig:micronet-teasor} summarizes the ImageNet performance, where MicroNets outperform the state-of-the-art by a large margin. In particular, our MicroNet models of 12M and 21M FLOPs outperform MobileNetV3 by 9.6\% and 4.5\% in terms of top-1 accuracy, respectively. For the extremely challenging regime of 6M FLOPs, MicroNet achieves 51.4\% top-1 accuracy, outperforming by 1.6\% over MobileNetV3, which is twice as complex (12M FLOPs). 

Even though MicroNet is manually designed for theoretical FLOPs, it outperforms MobileNetV3 (which is searched over inference latency) with fast inference on edge devices. Furthermore, our MicroNet surpasses MobileNetV3 on object detection and keypoint detection, but uses substantially less computational cost.


\section{Related Work}

\noindent \textbf{Efficient CNNs:}
MobileNets \cite{howard2017mobilenets, sandler2018mobilenetv2, Howard_2019_ICCV_mbnetv3} decompose $k \times k$ convolution into a depthwise and a pointwise convolution. ShuffleNets \cite{Zhang_2018_CVPR, ma_2018_ECCV} further simplify pointwise convolution by group convolution and channel shuffle. 
\cite{Tan-bmvc2019-mixconv} uses MixConv to mix up multiple kernel sizes in a convolution.
\cite{vahid_2020_CVPR} uses butterfly transform to approximate pointwise convolution. EfficientNet \cite{tan-ICML19-efficientnet, Tan_2020_CVPR} proposes a compound scaling method to scale depth/width/resolution uniformly.  AdderNet \cite{Chen_2020_CVPR_addernet} trades massive multiplications for cheaper additions. GhostNet \cite{Han_2020_CVPR_ghostnet} generates more feature maps from cheap linear transformations. Sandglass \cite{Daquan_2020_ECCV_RethinkingBS} alleviates information loss by flipping the structure of inverted residual block.
\cite{yu2018slimmable, Cai2019OnceFA} train one network to support multiple sub-networks.

\noindent \textbf{Dynamic Neural Networks:}
Dynamic networks improve the representation capability by adapting architectures or parameters to the input. 
\cite{NIPS2017_6813,liu2018ddnn,Wang_2018_ECCV,Wu_2018_CVPR} perform dynamic routing within a super-network. \cite{Wang_2018_ECCV} and \cite{Wu_2018_CVPR} use reinforcement learning to learn a controller for skipping part of an existing model. MSDNet \cite{huang2018multiscale} allows early-exit for easy samples based on the prediction confidence. \cite{Yuan2019S2DNASTS} searches for the optimal MSDNet. \cite{Li_2020_CVPR_dynamic_routing} learns dynamic routing across scales for semantic segmentation. \cite{Yang_2020_CVPR} adapts image resolution to achieve efficient inference.
Another line of work keeps the architectures fixed, but adapts parameters. HyperNet \cite{Ha2017HyperNetworks} uses another network to generate parameters for the main network. SENet \cite{Hu_2018_CVPR} adapt weights over channels based on squeezing global context. SKNet \cite{Li_2019_CVPR_SKNet} adapts attention over kernels with different sizes. Dynamic convolution \cite{Yang2019CondConvCP,Chen2019DynamicCA} aggregates multiple convolution kernels based on their attention. Dynamic ReLU \cite{Chen2020DynamicReLU} adapts slopes and intercepts of two linear functions in ReLU \cite{NairH10Relu,JarrettKRL09Relu}. 
\cite{Ma_2020_eccv_WeightNetRT} uses grouped fully connected layer to generate convolutional weights directly. 
\cite{Chen2020DynamicRC} presents spatial-aware dynamic convolution. 
\cite{Su_2020_eccv_DynamicGC} proposes dynamic group convolution. 
\cite{Tian_2020_eccv_ConditionalCF} applies dynamic convolution on instance segmentation.

\section{Micro-Factorized Convolution}
The goal of Micro-Factorized convolution is to optimize the trade-off between the number of channels and node connectivity. Here, the \textit{connectivity} $E$ of a layer is defined as the number of paths per output node, where a path connects an input node and an output node. 

\subsection{Micro-Factorized Pointwise Convolution} \label{micro-fac-point-conv}
\begin{figure*}[t]
	\begin{center}
		\includegraphics[width=1.0\linewidth]{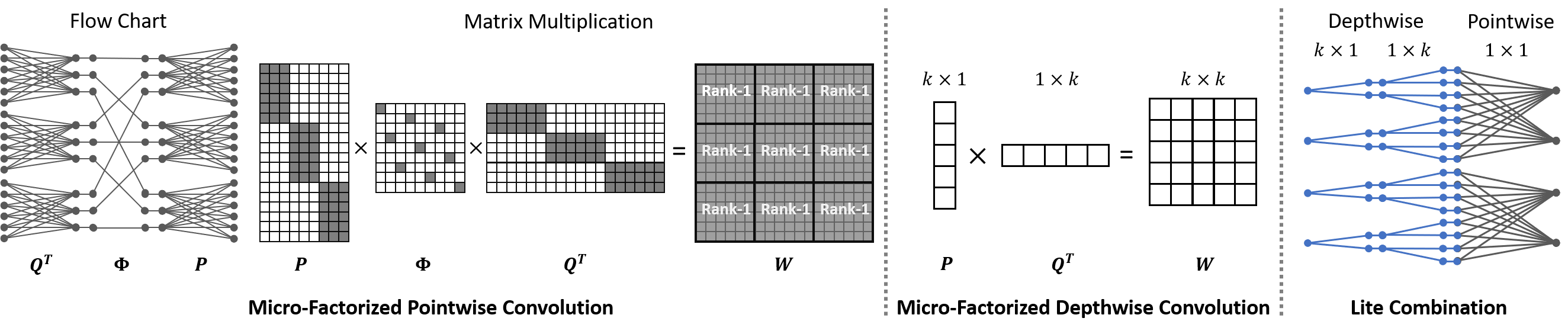}
	\end{center}
	\vspace{-3mm}
	\caption{\textbf{Micro-Factorized pointwise and depthwise convolutions}. \textbf{Left:} factorizing a pointwise convolution into two group-adaptive convolutions, where the group number $G=\sqrt{C/R}=\sqrt{18/2}=3$. The resulting matrix $\bm{W}$ can be divided into $G \times G$ blocks, of which each block has rank 1. \textbf{Middle:} factorizing a $k \times k$ depthwise convolution into a $k \times 1$ and a $1 \times k$ depthwise convolutions. \textbf{Right:} lite combination of Micro-Factorized pointwise and depthwise convolutions.}
	\label{fig:low-rank}
	\vspace{-2mm}
\end{figure*}
We propose the use of group-adaptive convolution to factorize a pointwise convolution. For conciseness, we assume the convolution kernel $\bm{W}$ has the same number of input and output channels ($C_{in} = C_{out} = C$) and ignore bias terms. The kernel matrix $\bm{W}$ is factorized into two group-adaptive convolutions, where the number of groups $G$ depends on the number of channels $C$, according to
\begin{align}
\bm{W} = \bm{P}\bm{\Phi}\bm{Q}^T,
\label{eq:mat-fac}
\end{align}
where $\bm{W}$ is a $C \times C$ matrix, $\bm{Q}$ is a  $C \times \frac{C}{R}$ matrix that compresses the number of channels by a factor of $R$, and $\bm{P}$ is a $C \times \frac{C}{R}$ matrix that expands the number of channels back to $C$.
$\bm{P}$ and $\bm{Q}$ are diagonal block matrices with $G$ blocks, each implementing the convolution of a group of channels. $\bm{\Phi}$ is a $\frac{C}{R} \times \frac{C}{R}$ permutation matrix, shuffling channels similarly to \cite{Zhang_2018_CVPR}. The computational complexity of the factorized layer is $\mathcal{O}=\frac{2C^2}{RG}$.
Figure \ref{fig:low-rank}-Left shows an example of the factorization, for $C=18$, $R=2$ and $G=3$.

\begin{figure}[t]
	\begin{center}
		\includegraphics[width=0.7\linewidth]{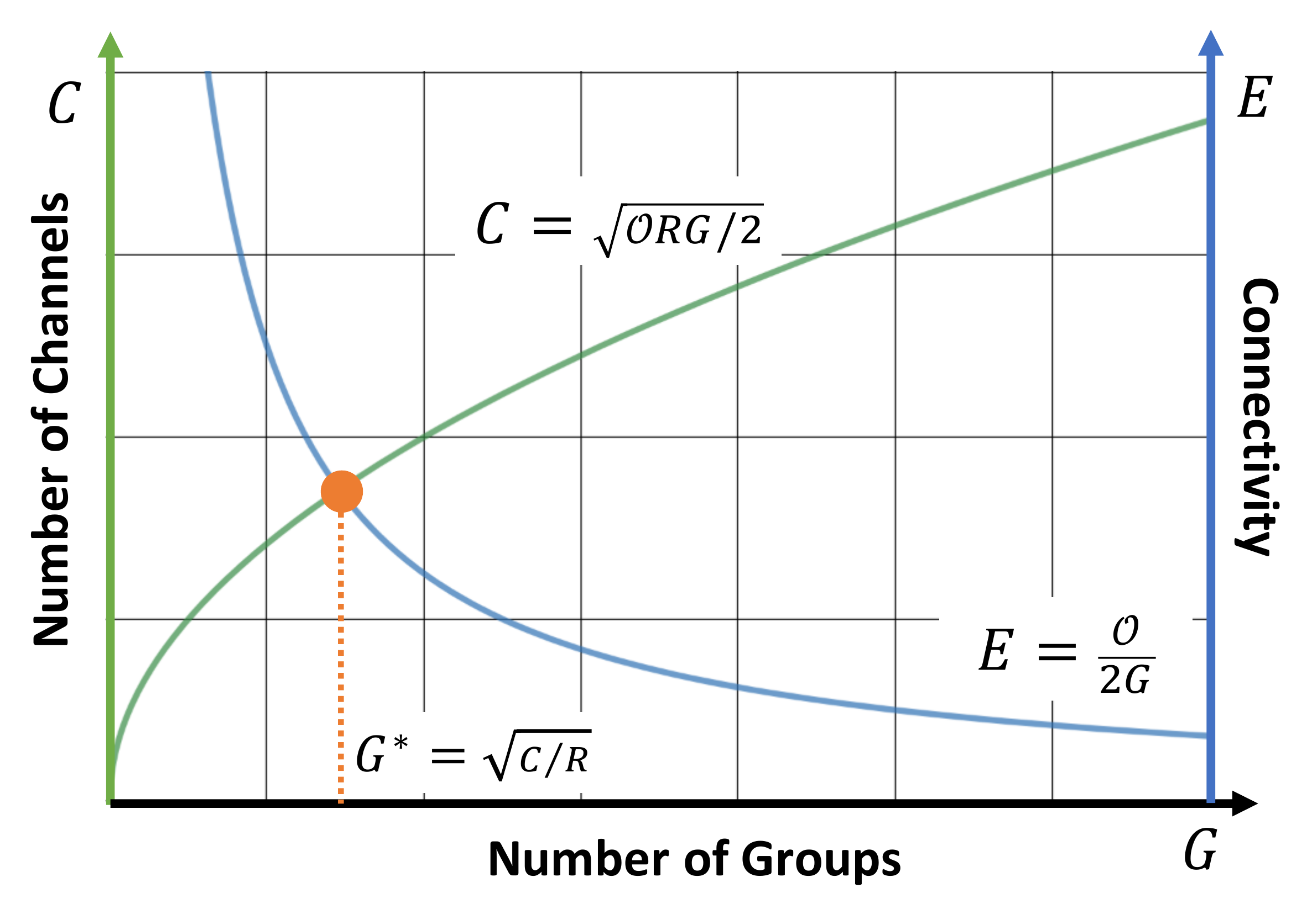}
	\end{center}
	\vspace{-2mm}
	\caption{\textbf{Number of Channels $C$ vs. Connectivity $E$} over number of groups $G$. We assume that the computational cost $\mathcal{O}$ and the reduction ratio $R$ are fixed. Best viewed in color.}
	\label{fig:hg2}
	\vspace{-2mm}
\end{figure}

The $\frac{C}{R}$ channels of matrix $\bm{\Phi}$ are denoted {\it hidden channels\/}. The grouping structure limits the number of these channels that are affected by (affect) each input (output) of the layer. Specifically, each hidden channel connects to $\frac{C}{G}$ input channels and each output channel connects to $\frac{C}{RG}$ hidden channels. The number $E=\frac{C^2}{RG^2}$ of input-output connections per output channel denotes the \textit{connectivity} $E$ of the layer.  When the computational budget $\mathcal{O}=\frac{2C^2}{RG}$ and the compression factor $R$ are fixed, the number of channels $C$ and connectivity $E$ change  with $G$ in opposite directions,
\begin{align}
C=\sqrt{\frac{\mathcal{O}RG}{2}},\;\; E = \frac{\mathcal{O}}{2G}.
\label{eq:width-connectivity}
\end{align}
This is illustrated in Figure \ref{fig:hg2}. As the number of groups $G$ increases, $C$ increases but $E$ decreases.  The two curves intersect ($C=E$) when 
\begin{align}
G=\sqrt{C/R},
\label{eq:g-cr}
\end{align}
in which case  
each output channel connects to all input channels exactly once ($E = C$). This guarantees that no redundant paths exist between any input-output pair (minimum path redundancy) while guaranteeing the existence of a path between each pair (maximum input coverage). Eq. \ref{eq:g-cr} is a defining property of micro-factorized pointwise convolution. It implies that the number of groups $G$ is \textit{not} fixed, but defined by the number of channels $C$ and the compression factor $R$, according to a square root law that optimally balances the number of channels $C$ and input/output connectivity. Mathematically, the resulting convolution matrix $\bm{W}$ is divided into $G \times G$ rank-1 blocks, as shown in Figure \ref{fig:low-rank}-Left.

\subsection{Micro-Factorized Depthwise Convolution} 
Figure \ref{fig:low-rank}-Middle shows how micro-factorization can be applied to
a $k\times k$ depthwise convolution. The convolution kernel is factorized into a $k\times1$ and a $1\times k$ kernel. 
This follows Eq. \ref{eq:mat-fac}, with per channel $k \times k$ kernel matrix $\bm{W}$, $k \times 1$ vector $\bm{P}$, $1 \times k$ vector $\bm{Q^T}$ and $\bm{\Phi}$ a scalar of value 1. 
This low rank approximation reduces the computational complexity from $\mathcal{O}(k^2C)$ to $\mathcal{O}(kC)$.

\vspace{1mm}
\noindent \textbf{Combining Micro-Factorized Pointwise and Depthwise Convolutions:}
Micro-Factorized pointwise and depthwise convolutions can be combined in two different ways: (a) regular combination, and (b) lite combination. The former simply concatenates the two convolutions. The lite combination, shown in Figure \ref{fig:low-rank}-Right, uses Micro-Factorized depthwise convolutions to expand the number of channels, by applying multiple spatial filters per channel. It then applies one group-adaptive convolution to fuse and squeeze the number of channels.
Compared to its regular counterpart, it spends more resources on learning spatial filters (depthwise) by saving channel fusion (pointwise) computations, which is empirically validated to be more effective for implementation of lower network layers.

\section{Dynamic Shift-Max}
So far, we have discussed the design of efficient static networks, which do not change their weights according to the input. We now introduce dynamic Shift-Max (DY-Shift-Max), a new dynamic non-linearity that strengthens connections between the groups created by micro-factorization. This is complementary to Micro-Factorized pointwise convolution, which focuses on connections within a group.

Let $\bm{x}=\{x_i\}$ ($i=1,\dots,C$) denote an input vector (or tensor) with $C$ channels that are divided into $G$ groups of $\frac{C}{G}$ channels each. The $j$-group circular shift (shifting $j\frac{C}{G}$ channels) of $\bm{x}$ is the vector $\bm{\hat{x}}^j$ such that $\hat{x}^j_i=x_{(i+j\frac{C}{G}) \bmod C}$. 
Dynamic Shift-Max outputs the maximum of $K$ fusions, each of which combines multiple ($J$) group shifts as:
\begin{align}
y_i &= \max_{1 \leq k \leq K}\{ \sum_{j=0}^{J-1} a^k_{i, j}(\bm{x})x_{(i+j\frac{C}{G}) \bmod C} \},
\label{eq:dynamic-group-shift-max}
\end{align}
where $a^k_{i,j}(\bm{x})$ is a dynamic weight, i.e. a weight that depends on the input $\bm{x}$. 
It is implemented as a hyper-function (with $CJK$ output dimension) that consists of a sequence of average pooling, two fully connected layers, and a sigmoid layer, as in Squeeze-and-Excitation \cite{squeezenet16}.

In this way, DY-Shift-Max implements two forms of non-linearity: it (a)  outputs the maximum of $K$ fusions of $J$ groups, and (b) weighs each fusion by a dynamic parameter $a^k_{i,j}(\bm{x})$. The first non-linearity is complementary to Micro-Factorized pointwise convolution, which focuses on connectivity within each group, strengthening the connections between groups. The second enables the network to tailor this strengthening to the input $\bm{x}$. The two operations increase the representation power of the network, compensating for the loss inherent to the reduced number of layers.  

DY-Shift-Max synthesizes $CJK$ weights $a^k_{i,j}(\bm{x})$ from input $\bm{x}$. Its computational complexity is a sum of (a) average pooling $\mathcal{O}(HWC)$, (b) generation of the $a^k_{i,j}(\bm{x})$ weights $\mathcal{O}(C^2JK)$, and (c) application of dynamic Shift-Max per channel and spatial location $\mathcal{O}(HWCJK)$. This leads to a light-weight model when $J$ and $K$ are small. Empirically, a good trade-off between classification performance and complexity is achieved when $J=2$ and $K=2$.


\begin{table*}[t]
	\begin{center}
	    \footnotesize
	    \setlength{\tabcolsep}{1.9mm}{
		\begin{tabular}{@{\hskip 1mm}c|c@{\hskip 2.2mm}c@{\hskip 2.2mm}c@{\hskip 2.2mm}c|c@{\hskip 2.2mm}c@{\hskip 2.2mm}c@{\hskip 2.2mm}c|c@{\hskip 2.2mm}c@{\hskip 2.2mm}c@{\hskip 2.2mm}c|c@{\hskip 2.2mm}c@{\hskip 2.2mm}c@{\hskip 2.2mm}c@{\hskip 1mm}}
		    \specialrule{.1em}{.05em}{.05em} 
			& \multicolumn{4}{c|}{M0} & \multicolumn{4}{c|}{M1} & \multicolumn{4}{c|}{M2} & \multicolumn{4}{c}{M3}   \\
			\cline{2-17}
			Output & Block & $k$ & $C$ & $\frac{C}{R}$ & Block & $k$ & $C$ & $\frac{C}{R}$ & Block & $k$ & $C$ & $\frac{C}{R}$ & Block & $k$ & $C$ & $\frac{C}{R}$\\
		
			\specialrule{.1em}{.05em}{.05em} 
			112$\times$112 &  stem & 3 & 4 & 2 &  stem & 3 & 6 & 3 &  stem & 3 & 8 & 4 &   stem & 3 & 12 & 4  \\
			\hline
			56$\times$56 & Micro-A & 3 & 16 & 8 &  Micro-A & 3 & 24 & 8 &  Micro-A & 3 & 32 & 12  &  Micro-A & 3 & 48 & 16  \\
			\hline
			&  Micro-A & 3 & 32 & 12 &  Micro-A & 3 & 32 & 16 &  Micro-A & 3 & 48 & 16 &  Micro-A & 3 & 64 & 24 \\
			28$\times$28 &   &  &  &    & & & & & Micro-B & 3 & 144 & 24 & Micro-B & 3 & 144 & 24   \\

			\hline
			& Micro-B & 5 & 64 & 16 & Micro-B & 5 & 96 & 16 & Micro-C  & 5 & 192 & 32 &  Micro-C & 3 & 192 & 32  \\
			& Micro-C & 5 & 128 & 32 & Micro-C & 5 & 192 & 32 & Micro-C  & 5 & 192 & 32 &  Micro-C & 5 & 192 & 32  \\
			14$\times$14&  &  &  &  &  &  &  &  & Micro-C & 5 & 384 & 64 & Micro-C & 5 & 384 & 64 \\
			&  &  &  &  &  &   &  &  & &  &  &  &Micro-C & 5 & 480 & 80   \\
			&  &  &  &  &  &   &  &  & &  &  &  &Micro-C & 5 & 480 & 80   \\
			\hline
			& Micro-C & 5 & 256 & 64 & Micro-C & 5 & 384 & 64 &  Micro-C & 5 & 576 & 96 & Micro-C  & 5 & 720 & 120  \\
			7$\times$7& Micro-C & 3 & 384 & 96 & Micro-C & 3 & 576 & 96 & Micro-C & 3 & 768 & 128 & Micro-C  & 3 & 720 & 120  \\
			&  &  &  &  &  &  & &  & &  & &  &  Micro-C & 3 & 864 & 144    \\
			\hline
			1$\times$1&  \multicolumn{16}{c}{avg pool $\rightarrow$ 2fc $\rightarrow$ softmax}\\
			\hline
			 & \multicolumn{4}{c|}{4M MAdds, 1.0M Param} &  \multicolumn{4}{c|}{6M MAdds, 1.8M Param} & \multicolumn{4}{c|}{12M MAdds, 2.4M Param} & \multicolumn{4}{c}{21M MAdds, 2.6M Param}  \\
			\specialrule{.1em}{.05em}{.05em} 
		\end{tabular}
		}
	\end{center}
	\vspace{-2mm}
	\caption{\textbf{MicroNet Architectures}. ``stem" refers to the stem layer. ``Micro-A", ``Micro-B", and ``Micro-C" refers to three Micro-Blocks (see section \ref{section:micro-blocks} and Figure \ref{fig:micro-block} for more details). $k$ is the kernel size, $C$ is the number of output channels, $R$ is the channel reduction ratio in Micro-Factorized pointwise convolution. Note that for ``Micro-A" (see Figure \ref{fig:micro-block}a), $C$ is the number of output channels in Micro-Factorized depthwise convolution, $\frac{C}{R}$ is the number of output channels for the block.}
	\vspace{-2mm}
	\label{table:micro-arch}
\end{table*}

\section{MicroNet} \label{section:micro-arch}
Below we describe in detail the design of MicroNet, using Micro-Factorized convolution and dynamic Shift-Max. 

\begin{figure}[t]
	\begin{center}
		\includegraphics[width=1.0\linewidth]{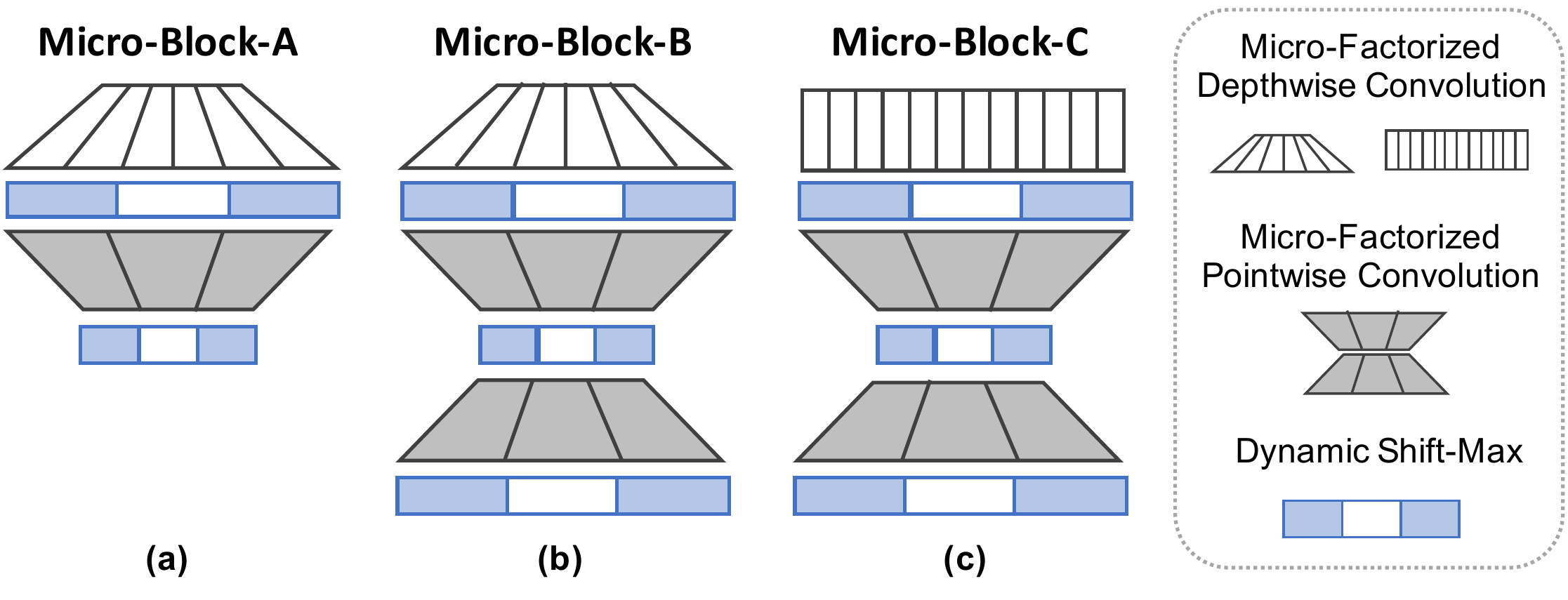}
	\end{center}
	\vspace{-2mm}
	\caption{\textbf{Diagram of three Micro-Blocks.} \textbf{(a) Micro-Block-A} that uses the lite combination of Micro-Factorized pointwise and depthwise convolutions (see Figure \ref{fig:low-rank}-Right). \textbf{(b) Micro-Block-B} that connects Micro-Block-A and Micro-Block-C. \textbf{(c) Micro-Block-C} that uses the regular combination of Micro-Factorized pointwise and depthwise convolutions. See Table \ref{table:micro-arch} for their usage.}
	\label{fig:micro-block}
	\vspace{-2mm}
\end{figure}

\subsection{Micro-Blocks} \label{section:micro-blocks}
MicroNet models consist of three Micro-Blocks of Figure \ref{fig:micro-block}, which combine Micro-Factorized pointwise and depthwise convolutions in different ways. All of the Micro-Blocks use the dynamic Shift-Max activation function.

\vspace{1mm}
\noindent \textbf{Micro-Block-A:} The Micro-Block-A of Figure \ref{fig:micro-block}a, uses the lite combination of Micro-Factorized pointwise and depthwise convolutions of Figure \ref{fig:low-rank}-Right.  It expands the number of channels with Micro-Factorized depthwise convolution, and compresses them with a group-adaptive convolution. It is best suited to implement lower network layers of higher resolution (e.g. $112 \times 112$ or $56 \times 56$).

\vspace{1mm}
\noindent \textbf{Micro-Block-B:} 
The Micro-Block-B of Figure \ref{fig:micro-block}b is used to connect Micro-Block-A and Micro-Block-C. Different from Micro-Block-A, it uses a full Micro-Factorized pointwise convolution, which includes two group-adaptive convolutions. Hence, it both compresses and expands the number of channels. All MicroNet models have a single Micro-Block-B (see Table \ref{table:micro-arch}). 

\vspace{1mm}
\noindent \textbf{Micro-Block-C:} 
The Micro-Block-C of Figure \ref{fig:micro-block}c implements the regular combination of Micro-Factorized depthwise and pointwise convolutions. It is best suited for the higher network layers (see Table \ref{table:micro-arch}) since it assigns more computation to channel fusion (pointwise) than the lite combination. The skip connection is used when the input and output have the same dimension.

Each micro-block has three hyper-parameters: kernel size $k$, number of output channels $C$, compression factor $R$ of the bottleneck of Micro-Factorized pointwise convolution. Note that the number of groups in the two group-adaptive convolutions is determined by Eq. \ref{eq:g-cr}. 

\subsection{Architectures}
All models are manually designed to optimize for FLOPs, which is a theoretical and device independent metric. We hope this can be leveraged by new hardware design and optimization for edge devices. We aware that FLOPs is \textit{not} equivalent to inference latency at existing hardware and will show in experiment that MicroNet also improves accuracy and latency.
We propose four models (M0, M1, M2, M3) of different computational cost (4M, 6M, 12M, 21M MAdds) based on the Micro-Blocks above. Table \ref{table:micro-arch} presents their full specification. These networks follow the same pattern from low to high layers: stem layer $\rightarrow$ Micro-Block-A $\rightarrow$ Micro-Block-B $\rightarrow$ Micro-Block-C. All models are handcrafted, without network architecture search (NAS). The network hyper-parameters are selected based on simple rules: $R$ is fixed (4 for M0, 6 for MicroNet-M1,M2,M3), $C$ increases from low to high levels, depth increases from M0 to M3. For the deepest model (M3), we only use one dynamic Shift-Max layer per block after the depthwise convolution. 
The stem layer includes a $3 \times 1$ convolution and a $1 \times 3$ group convolution, and is followed by a ReLU. The second convolution expands the number of channels. 

\subsection{Relation to Prior Work} 
MicroNet has various connections to the recent deep learning literature. It is related to the popular MobileNet \cite{howard2017mobilenets, sandler2018mobilenetv2, Howard_2019_ICCV_mbnetv3} and ShuffleNet \cite{Zhang_2018_CVPR, ma_2018_ECCV} models. It shares the inverted bottleneck structure with MobileNet and the use of group convolution with ShuffleNet. In contrast, MicroNet differs from these models in both its convolutions and activation functions. First, it factorizes pointwise convolutions into group-adaptive convolutions, with the number of groups $G=\sqrt{C/R}$ that is channel adaptive and guarantees minimum path redundancy. Second, it factorizes depthwise convolution. Third, it relies on a novel activation function, dynamic Shift-Max, to strengthen group connectivity in a non-linear and input dependent manner. Dynamic Shift-Max itself generalizes the recently proposed dynamic ReLU \cite{Chen2020DynamicReLU} (i.e. dynamic ReLU is a special case where $J=1$ and each channel is activated alone).

\section{Experiments}
We evaluate MicroNet on three tasks: (a) image classification, (b) object detection, and (c) keypoint detection. In this section, the baseline MobileNetV3-Small in \cite{Howard_2019_ICCV_mbnetv3} is denoted as MobileNetV3, for conciseness. 

\subsection{ImageNet Classification}
We start by evaluating the four MicroNet models (M0--M3) on the task of ImageNet \cite{deng2009imagenet} classification. ImageNet has 1000 classes, including 1,281,167 images for training and 50,000 images for validation. 

All models are trained using an SGD optimizer with 0.9 momentum. The image resolution is 224$\times$224. Data augmentation of standard random cropping and flipping is used. We use a mini-batch size of 512, and a learning rate of 0.02. Each model is trained for 600 epochs with cosine learning rate decay. The weight decay is 3e-5 and dropout rate is $0.05$ for smaller MicroNets (M0, M1, M2). For the largest model M3, the weight decay is 4e-5 and dropout rate is $0.1$. 

\begin{table}[t!]
	\begin{center}
	    \footnotesize
	    \setlength{\tabcolsep}{1.7mm}{
		\begin{tabular}{c@{\hskip 2mm}|@{\hskip 1mm}c@{\hskip 2mm}c@{\hskip 2mm}c@{\hskip 1mm}|@{\hskip 1mm}c@{\hskip 2mm}c@{\hskip 1mm}|@{\hskip 1mm}c@{\hskip 2mm}c@{\hskip 1mm}|@{\hskip 1mm}c}
		    \specialrule{.1em}{.05em}{.05em}
		    & \multicolumn{3}{@{\hskip 1mm}c|@{\hskip 1mm}}{Micro-Fac Conv} & \multicolumn{2}{@{\hskip 1mm}c|@{\hskip 1mm}}{Shift-Max} & & & \\
		    & DW & PW & Lite & static & dynamic & Param & MAdds & Top-1 \\
		   \specialrule{.1em}{.05em}{.05em}
		    Mobile & & & & & &1.3M &10.6M &44.9 \\
		    \hline
		    &\checkmark  &  & & & & 1.7M & 10.6M & 46.4\\
		    &\checkmark &\checkmark & & & &1.7M & 10.6M &50.0 \\
            Micro&\checkmark & \checkmark & \checkmark & & & 1.8M & 10.5M & 51.7 \\
            & \checkmark & \checkmark & \checkmark & \checkmark & & 1.9M & 11.8M & 54.4 \\
            & \checkmark & \checkmark & \checkmark &  & \checkmark & 2.4M & 12.4M & \textbf{58.5} \\
	       \specialrule{.1em}{.05em}{.05em}
		\end{tabular}
		}
	\end{center}
	\vspace{-2mm}
	\caption{\textbf{The path from MobileNet to MicroNet} evaluated on ImageNet classification. Here, we modify MobileNet-V2 such that it has similar FLOPs (about 10.6M) to three Micro-Factorized convolution options: depthwise (DW), pointwise (PW), and lite combination at low levels (Lite). We also compare dynamic Shift-Max with its static counterpart (static $a^k_{i,j}$ in Eq. \ref{eq:dynamic-group-shift-max}).
	}
	\vspace{-2mm}
	\label{table:mobile-to-micro}
\end{table}

\vspace{-2mm}
\subsubsection{Ablation Studies}
Several ablations were performed using MicroNet-M2. All models are trained for 300 epochs. The default hyper parameters of DY-Shift-Max were set as $J$=2, $K$=2.

\vspace{1mm}
\noindent \textbf{From MobileNet to MicroNet:} 
Table \ref{table:mobile-to-micro} shows the path from MobileNet to MicroNet. Both share the inverted bottleneck structure. Here, we modify MobileNetV2 
(without SE \cite{Hu_2018_CVPR}) such that it has complexity (10.6M MAdds) similar to the static Micro-Factorized convolution variants of row 2--4. The introduction of Micro-Factorized depthwise convolutions improves performance by $1.5\%$. Micro-Factorized pointwise convolutions adds another $3.6\%$ and the lite combination at lower layers adds a final gain of $1.7\%$. Altogether the three factorizations boost the top-1 accuracy of the static network from 44.9\% to 51.7\%.
The addition of static and dynamic Shift-Max further increases this gain by 2.7\% and 6.8\% respectively, for a small increase in computation. This demonstrates that both \textit{Micro-Factorized Convolutions} and \textit{Dynamic Shift-Max} are effective and complementary mechanisms for the implementation of networks with extremely low computational cost.

\vspace{1mm}
\noindent \textbf{Number of Groups $G$:}
Micro-Factorized pointwise convolution includes two group-adaptive convolutions, with a number of groups equal to the integer closest to $G=\sqrt{C/R}$. Table \ref{table:mfc-pointwise}a compares this to networks of similar structure and FLOPs (about 10.5M MAdds), but using a fixed group cardinality. Group-adaptive convolution achieves higher accuracy, demonstrating the importance of its optimal trade-off between input/output connectivity and the number of channels.

This is further confirmed by Table \ref{table:mfc-pointwise}b, which compares different options for the adaptive number of groups. This is controlled by a multiplier $\lambda$ such that $G=\lambda \sqrt{C/R}$. Larger $\lambda$ corresponds to more channels but less input/output connectivity (see Figure \ref{fig:hg2}). The optimal balance is achieved when $\lambda$ is between 0.5 and 1. Top-1 accuracy drops when $\lambda$ either increases (more channels but less connectivity) or decreases (fewer channels but more connectivity) from this optimal point. The value $\lambda=1$ is used in the remainder of the paper. Note that all models in Table \ref{table:mfc-pointwise}b have similar computational cost (about 10.5M MAdds). 

\begin{table*}[t]
\parbox{.33\linewidth}{
    \begin{center}
	    \footnotesize
		\begin{tabular}{c@{\hskip 2.5mm}|c@{\hskip 2.5mm}c@{\hskip 2.5mm}c@{\hskip 2.5mm}}
		    \specialrule{.1em}{.05em}{.05em} 
			$G$ & Param & MAdds  & Top-1  \\[0.5em]
			\specialrule{.1em}{.05em}{.05em} 
			 1 & 1.3M & 10.6M  & 48.8  \\
			 2 & 1.5M & 10.5M  & 50.2 \\
			 4 & 1.7M & 10.6M & 50.7 \\
			 8 & 1.7M & 10.6M & 50.8 \\
			\hline
            $G=\sqrt{C/R}$ & 1.8M & 10.5M  & \textbf{51.7} \\
			\specialrule{.1em}{.05em}{.05em} 
			\multicolumn{4}{c}{} \\[-0.5em]
			\multicolumn{4}{c}{(a) \textbf{Fixed group number $G$.}} \\
		\end{tabular}
	\end{center}
	\label{table:group-number-1}
	\vspace{-0.5em}
}
\hfill
\parbox{.33\linewidth}{
    \begin{center}
	    \footnotesize
		\begin{tabular}{c@{\hskip 2.5mm}|r@{\hskip 2.5mm}r@{\hskip 2.5mm}r@{\hskip 2.5mm}}
		    \specialrule{.1em}{.05em}{.05em} 
			$\lambda=\frac{G}{\sqrt{C/R}}$ & Param & MAdds  & Top-1  \\
			\specialrule{.1em}{.05em}{.05em} 
			$\;\;\;$ $0.25$ & 1.5M & 10.5M   & 50.2    \\
			$\;\;\;\;$ $0.5$ & 1.7M & 10.6M  & 51.6 \\
            \xmark$\:$ $1.0$ & 1.8M & 10.5M  & \textbf{51.7} \\
            $\;\;\;\;$ $2.0$ & 2.1M & 10.5M  & 50.6 \\
            $\;\;\;\;$ $4.0$ & 2.2M & 10.7M & 47.6 \\
			\specialrule{.1em}{.05em}{.05em} 
			\multicolumn{4}{c}{}\\[-0.5em]
			\multicolumn{4}{c}{(b) \textbf{Adaptive group number $G$.}} \\
		\end{tabular}
	\end{center}
	\label{table:group-number-2}
	\vspace{-0.5em}
}
\hfill
\parbox{.33\linewidth}{
    \begin{center}
	    \footnotesize
		\begin{tabular}{r@{\hskip 2.5mm}c|c@{\hskip 2.5mm}c@{\hskip 2.5mm}c}
		    \specialrule{.1em}{.05em}{.05em}
		    \multicolumn{2}{c|}{Levels} & & & \\
		    low & high & Param & MAdds  & Top-1  \\
		   \specialrule{.1em}{.05em}{.05em}
		     & &1.7M &10.6M &  50.0 \\
		     \xmark$\:$ \checkmark & & 1.8M &10.5M & \textbf{51.7} \\
		     \checkmark & \checkmark &2.0M & 10.6M & 51.2 \\
	       \specialrule{.1em}{.05em}{.05em}
	       \multicolumn{5}{c}{}\\
	       \multicolumn{5}{c}{}\\[-0.5em]
	       \multicolumn{5}{c}{}\\[-0.5em]
		\multicolumn{5}{c}{(c) \textbf{Lite combination} at different levels} \\
		\end{tabular}
	\end{center}
    \label{table:half-and-half}
	\vspace{-0.5em}
}
\caption{\textbf{Ablations of Micro-Factorized convolution} on ImageNet classification. 
\xmark $\:$ indicates the default choice for the rest of the paper.}
\label{table:mfc-pointwise}
\vspace{-2mm}
\end{table*}

\vspace{1mm}
\noindent \textbf{Lite combination:} 
Table \ref{table:mfc-pointwise}c compares using the lite combination of Micro-Factorized pointwise and depthwise convolutions (Figure \ref{fig:low-rank}-Right) at different layers. The lite combination is more effective for lower layers. Compared to the regular combination, it saves computations from channel fusion (pointwise) to allow more spatial filters (depthwise).

\vspace{1mm}
\noindent \textbf{Activation functions:} 
Dynamic Shift-Max is compared to three previous activation functions: ReLU \cite{NairH10Relu}, SE+ReLU \cite{Hu_2018_CVPR}, and dynamic ReLU \cite{Chen2020DynamicReLU}. Table \ref{table:ablation-dy-activation} shows that dynamic Shift-Max outperforms all three by a clear margin (at least 2.5\%). Note that dynamic ReLU is the special case of dynamic Shift-Max with $J=1$ (see Eq. \ref{eq:dynamic-group-shift-max}). 

\begin{table}[t!]
	\begin{center}
	    \footnotesize
	    \setlength{\tabcolsep}{3.1mm}{
		\begin{tabular}{l|rr|ll}
		    \specialrule{.1em}{.05em}{.05em} 
			Activation & Param & MAdds &Top-1 &	Top-5   \\
		
			\specialrule{.1em}{.05em}{.05em} 
			ReLU\cite{NairH10Relu} & 1.8M & 10.5M & 51.7  & 74.3 \\
			SE\cite{Hu_2018_CVPR}+ReLU & 2.1M & 10.9M & 54.4 & 76.8 \\
			Dynamic ReLU \cite{Chen2020DynamicReLU} & 2.4M & 11.8M & 56.0 & 78.0 \\
			\hline
			Dynamic Shift-Max & 2.4M & 12.4M & \textbf{58.5} & \textbf{80.1} \\
			\specialrule{.1em}{.05em}{.05em} 
		\end{tabular}
		}
	\end{center}
	\vspace{-2mm}
	\caption{\textbf{Dynamic Shift-Max vs. other activation functions} on ImageNet classification. MicroNet-M2 is used.}
	\vspace{-2mm}
	\label{table:ablation-dy-activation}
\end{table}

\begin{table}[t!]
	    \footnotesize
	    \setlength{\tabcolsep}{2.7mm}{
        \begin{tabular}{c|c@{\hskip 2.5mm}c@{\hskip 2.5mm}c|c@{\hskip 2.5mm}c|c@{\hskip 2.5mm}c}
        \specialrule{.1em}{.05em}{.05em}
         & $A_1$ & $A_2$ & $A_3$ & Param & MAdds & Top-1 & Top-5 \\
        \specialrule{.1em}{.05em}{.05em}
        ReLU & -- & -- & -- &1.8M & 10.5M&51.7 & 74.3 \\
        \hline
         & \checkmark & -- & -- & 2.1M& 11.3M&55.9 & 77.9 \\
         & -- & \checkmark & -- &2.0M  & 10.6M&53.3 & 76.0 \\
        Dynamic & -- & -- & \checkmark& 2.1M &11.2M & 54.8 & 77.2 \\
        Shift-Max & \checkmark & \checkmark & -- & 2.2M&11.5M &56.6& 78.3\\
         & \checkmark & -- & \checkmark & 2.3M &12.2M &57.9 & 79.6\\
         & -- & \checkmark & \checkmark &2.2M &11.4M &55.5 & 77.8 \\        
         & \checkmark & \checkmark & \checkmark &2.4M &12.4M & \textbf{58.5} & \textbf{80.1} \\
        \specialrule{.1em}{.05em}{.05em}
         \multicolumn{6}{c}{}\\
        \end{tabular}
        }
        \vspace{-2mm}
        \caption{\textbf{Dynamic Shift-Max at different layers} evaluated on ImageNet. MicroNet-M2 is used. $A_1,A_2,A_3$ indicate three activation layers sequentially in Micro-Block-B and Micro-Block-C (see Figure \ref{fig:micro-block}). Micro-Block-A only includes $A_1$ and $A_2$.}
        \label{table:ablation-diff-layer}
	\vspace{-2mm}
\end{table}

\vspace{1mm}
\noindent \textbf{Location of DY-Shift-Max:}
Table \ref{table:ablation-diff-layer} shows the top-1 accuracy when dynamic Shift-Max is implemented in different combinations of the three layers of the micro-blocks of Figure \ref{fig:micro-block}. When used in a single layer, dynamic Shift-Max should be placed after the depthwise convolution. This improves the top-1 accuracy over a network with ReLU activations by $4.2\%$. Adding a Dynamic Shift-Max activation at the Micro-Block output further improves performance by $2\%$. Finally, using three layers of Dynamic Shift-Max further increases the gain over the ReLU network to $6.8\%$.


\begin{table}[t!]
	\begin{center}
	    \footnotesize
	    \setlength{\tabcolsep}{3.7mm}{
		\begin{tabular}{rc|cc|cc}
		    \specialrule{.1em}{.05em}{.05em} 
			$J$ & $K$ & Param & MAdds &Top-1 &	Top-5   \\
		
			\specialrule{.1em}{.05em}{.05em} 
			1 & 1 & 2.1M & 10.9M & 54.4  & 76.8 \\
			 \hline
			2 & 1 & 2.2M & 11.8M & 55.9 & 78.2 \\
			\xmark$\:$ 2 & 2 & 2.4M & 12.4M & 58.5 & 80.1 \\
			2 & 3 & 2.6M & 13.8M& 58.1 & 79.7 \\
			\hline
			1 & 2 & 2.2M & 11.2M & 55.5 & 77.6 \\
			\xmark$\:$ 2 & 2 & 2.4M & 12.4M & 58.5 & 80.1 \\
			3 & 2 & 2.6M & 14.2M & 59.0 & \textbf{80.3} \\
			\hline
			3 & 3 & 2.8M & 15.3M & \textbf{59.1} & \textbf{80.3} \\
			\specialrule{.1em}{.05em}{.05em} 
		\end{tabular}
		}
	\end{center}
	\vspace{-2mm}
	\caption{\textbf{Ablations of two hyper parameters in dynamic Shift-Max} ($J$, $K$ in Eq. \ref{eq:dynamic-group-shift-max}) on ImageNet classification. \xmark $\:$ indicates the default choice for the rest of the paper.}
	\label{table:ablation-J-K}
	\vspace{-2mm}
\end{table}

\vspace{1mm}
\noindent \textbf{Hyper-parameters in DY-Shift-Max:} 
Table \ref{table:ablation-J-K} shows the results of using different combinations of $K$ and $J$ in Eq. \ref{eq:dynamic-group-shift-max}. We add a ReLU when $K=1$ as only one element is left in the max operator. The baseline of the first row ($J=1$, $K=1$) is equivalent to SE+ReLU \cite{Hu_2018_CVPR}. For fixed $J=2$ (fusion of two groups), the best of two fusions ($K=2$) is better than a single fusion ($K=1$), but adding a third fusion does not help, since it only adds path redundancy. When $K$ is fixed at $K=2$ (best of two fusions), fusing more groups $J$ is consistently better but requires more FLOPs. A good tradeoff is achieved with $J=2$ and $K=2$, enabling a gain of 4.1\% over the baseline, for an additional 1.5M MAdds.

\vspace{-2mm}
\subsubsection{Comparison to Prior Networks}
Table \ref{table:imagenet-cls-result} compares MicroNet to the state-of-the-art models, which have complexity less than 24M FLOPs. As the prior works lack of reported results within 10M FLOPs budget, we extend the popular MobileNetV3 to 6M and 4M FLOPs as baseline, by using width multiplier 0.2 and 0.15 respectively. They share the same training setup with MicroNet. 

To make comparison fair, two variations of M1--M3 (e.g. M3$^{\#}$ and M3) are used. The former (M3$^{\#}$) 
requires similar model size to but fewer FLOPs than the baseline (MobileNetV3 0.5$\times$). The latter (M3) requires similar FLOPs but allows more parameters (up to 1M), best serving scenarios that FLOPs is more critical than memory. This is due to the difficulty to match both model size and FLOPs, except for the smallest model (M0). Note that M3$^{\#}$ has similar structure to M3, only shrinking the model size by reducing network width and parameters in dynamic Shift-Max. 

In all cases, MicroNet outperforms all prior networks by a clear margin. For instance, MicroNet-M1$^{\#}$, M2$^{\#}$, M3$^{\#}$ outperform their MobileNetV3 counterpart by 8.3\%, 8.4\%, and 3.3\%, respectively. Given another 1M budget on model size, MicroNet-M1, M2, M3 increase these gains by 2.0\%, 1.2\% and 1.2\%, respectively. 
MicroNet-M0 outperforms MobileNetV3 0.15$\times$ by 12.9\% (46.6\% vs. 33.7\%), demonstrating its better handle of cutting computational cost from 6M to 4M MAdds. In particular, the top-1 accuracy drops by 4.8\% from MicroNet-M1 to M0, while the accuracy degrades by 7.4\% from MobileNetV3 $\times$0.2 to $\times$0.15.
When compared to recent MobileNet and ShuffleNet improvements, such as 
ButterflyTransforms \cite{vahid_2020_CVPR} and TinyNet \cite{NEURIPS2020_e069ea4c}, MicroNet models have gains of more than 2.6\% top-1 accuracy but use less FLOPs. This demonstrates the effectiveness of MicroNet at extremely low FLOPs.

\begin{table}[t!]
	\begin{center}
	    \footnotesize
	    \setlength{\tabcolsep}{0.9mm}{
		\begin{tabular}{l@{\hskip 2.5mm}|c@{\hskip 2.5mm}r@{\hskip 2.5mm}|c@{\hskip 2.5mm}c@{\hskip 2mm}}
		    \specialrule{.1em}{.05em}{.05em} 
			Model & \#Param & MAdds &Top-1 &	Top-5   \\
		
			\specialrule{.1em}{.05em}{.05em} 
			MobileNetV3 0.15$\times^{\dag}$ & 1.0M & 4M & 33.7 & 57.2\\
			\textbf{MicroNet-M0}  & 1.0M & 4M &  \textbf{46.6} & \textbf{70.6}  \\
			\hline
			MobileNetV3 0.2$\times^{\dag}$ & 1.2M & 6M & 41.1 & 65.2	  \\ 
			\textbf{MicroNet-M1$^{\#}$}  & 1.2M & 5M &  49.4 & 72.9  \\
			\textbf{MicroNet-M1}  & 1.8M & 6M &  \textbf{51.4} & \textbf{74.5}  \\
			
			\hline
			ShuffleNetV1 0.25$\times$ \cite{Zhang_2018_CVPR} & -- & 13M & 47.3 & -- \\
		    MobileNetV3 0.35$\times$ \cite{Howard_2019_ICCV_mbnetv3} & 1.4M & 12M  & 49.8 & -- 		 \\
		    HBONet (96$\times$96) \cite{li2019hbonet} & -- & 12M & 50.3 & 73.8 \\
		    MobileNetV3+BFT 0.5$\times$ \cite{vahid_2020_CVPR} & -- & 15M & 55.2 & -- \\
		    \textbf{MicroNet-M2$^{\#}$}  & 1.4M  & 11M & 58.2 & 80.1 		 \\
		    \textbf{MicroNet-M2}  & 2.4M  & 12M & \textbf{59.4} & \textbf{80.9} 		 \\
			\hline
			HBONet (128$\times$128) \cite{li2019hbonet} & -- & 21M & 55.2 & 78.0 \\
			ShuffleNetV2+BFT \cite{vahid_2020_CVPR} & -- & 21M & 57.8 & -- \\
			MobileNetV3 0.5$\times$ \cite{Howard_2019_ICCV_mbnetv3} & 1.6M & 21M & 58.0 & 	--	 \\
			TinyNet-E (106$\times$106) \cite{NEURIPS2020_e069ea4c} & 2.0M & 24M & 59.9 & 81.8 \\
			\textbf{MicroNet-M3$^{\#}$}  & 1.6M & 20M & 61.3 &  82.9	 \\
			\textbf{MicroNet-M3}  & 2.6M & 21M & \textbf{62.5} &  \textbf{83.1}		 \\
			\specialrule{.1em}{.05em}{.05em} 
		\end{tabular}
		}
	\end{center}
	\vspace{-1mm}
	\caption{\textbf{ImageNet \cite{deng2009imagenet} classification results}. $^{\#}$ stands for the MicroNet variation that has similar model size to but fewer MAdds than the corresponding MobileNetV3-Small baseline. $^{\dag}$ indicates our implementation under the same training setup with MicroNet. 
	``--": not available in the original paper. Note that input resolution 224$\times$224 is used for MicroNet and related works other than HBONet/TinyNet, whose input resolution is shown in the bracket.}
	\label{table:imagenet-cls-result}
\end{table}

\vspace{-2mm}
\subsubsection{Inference Latency}

\begin{figure}[t]
	\begin{center}
		\includegraphics[width=1.0\linewidth]{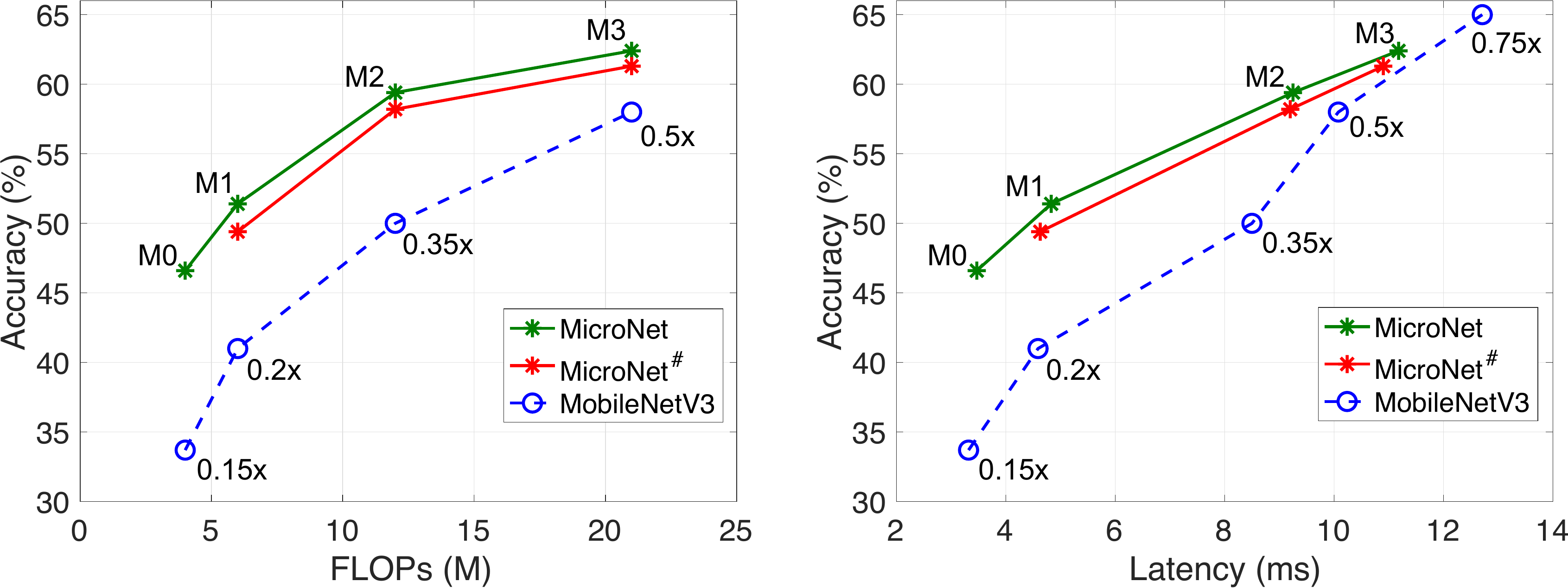}
	\end{center}
	\vspace{-2mm}
	\caption{Evaluation on ImageNet classification. \textbf{Left}: \textit{top-1 accuracy vs. FLOPs}. \textbf{Right}: \textit{top-1 accuracy vs. latency}. Note that MobileNetV3 $\times$0.75 is added to facilitate the comparison. MicroNet outperforms MobileNetV3, especially at extremely low computational cost (more than 5\% gain on top-1 accuracy when FLOPs is less than 15M or latency is less than 9ms).}
	\label{fig:latency}
	\vspace{-1.5mm}
\end{figure}

We also measure the inference latency of MicroNet on an Intel(R) Xeon(R) CPU E5-2620 v4 (2.10GHz). Following the common settings in \cite{sandler2018mobilenetv2, Howard_2019_ICCV_mbnetv3}, we test under single-threaded mode with batch size $1$. The average inference latency of 5,000 images (with resolution 224$\times$224) is reported. Figure \ref{fig:latency}-Right shows the comparison between MicroNet and MobileNetV3-Small. To achieve similar performance, MicroNet clearly consumes less runtime than MobileNetV3. For example, MicroNet with 55\% accuracy has a latency less than 7ms, while MobileNetV3 requires about 9.5ms. The accuracy-latency curve is slightly degraded when using MicroNet with fewer parameters (M1$^{\#}$, M2$^{\#}$, M3$^{\#}$), but it still outperforms MicroNetV3. 
Although the largest MicroNet model (M3) only slightly outperforms MobileNetV3 for the same latency, MicroNet gains significantly more improvement over MobileNetV3 when the latency decreases. In particular, at a latency of 4ms, MicroNet improves over MobileNetV3 by 10\%, demonstrating its strength at low computational cost.

\vspace{-2mm}
\subsubsection{Discussion}
As shown in Figure \ref{fig:latency}, MicroNet clearly outperforms MobileNetV3 under the same FLOPs, but the gap shrinks under the same latency. This is due to two reasons. First, different from MobileNetV3 that is optimized for latency by search, MicroNet is manually designed based on theoretical FLOPs. Second, the implementation of group convolution and dynamic Shift-Max are not optimized (we use PyTorch for implementation). We observe that the latency of group convolution is not proportionally reduced as the number of groups increases, and dynamic Shift-Max is significantly slower than convolution with the same FLOPs. 

We believe that the runtime performance of MicroNet can be further improved by using hardware-aware architecture search to find latency friendly combination of Micro-Factorized convolution and dynamic Shift-Max. MicroNet can also leverage the improvement of optimization in group convolution \cite{GroupOptimization-IEEE} and dynamic Shift-Max to speed up. We will investigate these in the future work.

\subsection{Object Detection}
We evaluate the generalization ability of MicroNet on COCO object detection \cite{lin2014microsoft}. All models are trained on \texttt{train2017} and evaluated in mean Average Precision (mAP) on \texttt{val2017}. Following \cite{Han_2020_CVPR_ghostnet}, MicroNet is used as a drop-in replacement for the backbone feature extractor in both the two-stage Faster R-CNN \cite{ren2015faster} with Feature Pyramid Networks (FPN) \cite{lin2017feature} and the one-stage RetinaNet \cite{lin2017focal}. All models are trained using SGD for 36 epochs (3$\times$) from ImageNet pretrained weights with the hyper-parameters and data augmentation suggested in \cite{wu2019detectron2}.

The detection results are shown in Table \ref{table:od-results}, where the backbone FLOPs are calculated using image size $224\times224$ as common practice. With significantly lower backbone FLOPs (21M \textit{vs} 56M), MicroNet-M3 achieves higher mAP than MobileNetV3-Small $\times$1.0 both on Faster R-CNN and RetinaNet frameworks, demonstrating its capability to transfer to detection task.  

\begin{table*}[t!]
    \begin{minipage}[b]{.37\linewidth}
      \centering
      \footnotesize
      \setlength{\tabcolsep}{0.9mm}{
        \begin{tabular}{c|c|cc}
		    \specialrule{.1em}{.05em}{.05em} 
			Backbone& DET Framework& MAdds & mAP   \\
		
			\specialrule{.1em}{.05em}{.05em} 
			MobileNetV3 $\times$1.0  &  & 56M & 25.9 \\
			\textbf{MicroNet-M3}  & R-CNN & 21M  &  \textbf{26.2} \\
			\textbf{MicroNet-M2}  &  & 12M  & 22.7  \\
			\specialrule{.1em}{.05em}{.05em} 
			MobileNetV3 $\times$1.0 &  & 56M & 24.0 \\
			\textbf{MicroNet-M3}  & RetinaNet  & 21M  &  \textbf{25.4} \\
			\textbf{MicroNet-M2}  & & 12M  & 22.6  \\
			\specialrule{.1em}{.05em}{.05em} 
			\multicolumn{4}{c}{}\\[-0.5em]
		\end{tabular}
	  }
	  \caption{\textbf{COCO object detection results}. All models are trained on \texttt{train2017} for 36 epochs ($3 \times$) and tested on \texttt{val2017}. MAdds is computed on image size 224$\times$224.}
	  \label{table:od-results}
    \end{minipage} 
    \quad
    \begin{minipage}[b]{.60\linewidth}
      \centering
        \footnotesize
        \setlength{\tabcolsep}{1.0mm}{
		\begin{tabular}{cc| r r| l c c c c}
			\specialrule{.1em}{.05em}{.05em} 
			Backbone & Head & Param & MAdds & AP &	AP$^{0.5}$ & AP$^{0.75}$ & AP$^M$ & AP$^L$\\
			\specialrule{.1em}{.05em}{.05em}
			MobileNetV3 $\times$1.0  &  Mobile-Blocks  & 2.1M & 726.9M & 57.1 & 83.8 & 63.7 &	55.0&		62.2\\	
			\hline
			\textbf{MicroNet-M3} & Micro-Blocks & 2.2M & 163.2M & \textbf{58.7} & \textbf{84.0} & \textbf{65.5}& \textbf{56.0} & \textbf{64.2} \\
			\textbf{MicroNet-M2} & Micro-Blocks & 1.8M & 116.8M & 54.9 & 82.0 & 60.3  & 53.2 & 59.6\\
			\specialrule{.1em}{.05em}{.05em} 
			\multicolumn{9}{c}{}\\[-0.5em]
		\end{tabular}
		}
		\caption{\textbf{COCO keypoint detection results}. All models are trained on \texttt{train2017} and tested on \texttt{val2017}. Input resolution 256$\times$192 is used. The baseline applies MobileNetV3-Small $\times$1.0 as backbone and the head structure in \cite{Chen2019DynamicCA} (which includes bilinear upsampling and inverted residual bottleneck blocks). Compared to the baseline, MicroNet-M3 has similar model size, consumes significantly less MAdds, but achieves higher accuracy.
	    }
	    \label{table:coco-kp}
    \end{minipage} 
    \vspace{-2mm}
\end{table*}

\subsection{Human Pose Estimation}

We also evaluate MicroNet on COCO single person keypoint detection. All models are trained on \texttt{train2017} that includes $57K$ images and $150K$ person instances labeled with 17 keypoints, and evaluated on \texttt{val2017} that contains 5000 images, using the mean average precision (AP) over 10 object key point similarity (OKS) thresholds.
Similar to object detection, two MicroNet models (M2, M3) are considered. The models are modified for the keypoint detection task, by increasing the resolution ($\times$2) of a select set of blocks (all blocks with stride of 32). Each model contains a head with three micro-blocks (one of stride 8 and two of stride 4) and a pointwise convolution that generates heatmaps for 17 keypoints. Bilinear upsampling is used to increase the head resolution, and the spatial attention mechanism of \cite{Chen2020DynamicReLU} is used.
Both models are trained from scratch for 250 epochs using Adam optimizer \cite{kingma:adam}. The human detection boxes are cropped and resized to 256$\times$192. The training and testing follow the setup of \cite{xiao2018simplebaseline, sun2019deep}.

Table \ref{table:coco-kp} compares MicroNet-M3 and M2 with a strong efficient baseline, which only requires 726.9M MAdds and 2.1M parameters. The baseline applies MobileNetV3-Small $\times$1.0 as backbone and mobile blocks (inverted residual bottleneck blocks) in the head (see \cite{Chen2019DynamicCA} for details). 
Our MicroNet-M3 only consumes 22\% (163.2M/726.9M) of the FLOPs used by the baseline but achieves higher performance, demonstrating its effectiveness for low-complexity keypoint detection. MicroNet-M2 provides a good handle for even lower complexity (116.8M FLOPs).

\section{Conclusion}
In this paper, we have presented MicroNet to handle extremely low computational cost. It builds on two proposed operators: Micro-Factorized convolution and Dynamic Shift-Max. The former balances between the number of channels and input/output connectivity via low rank approximations on both pointwise and depthwise convolutions. The latter fuses consecutive channel groups dynamically, enhancing both node connectivity and non-linearity to compensate for the depth reduction. A family of MicroNets achieve solid improvement for three tasks (image classification, object detection and human pose estimation) under extremely low FLOPs. We hope this work provides good baselines for efficient CNNs on multiple vision tasks.

\section*{Acknowledgement}
This work was partially funded by NSF awards IIS-1924937, IIS-2041009.

{\small
\bibliographystyle{ieee_fullname}
\bibliography{egbib}
}

\end{document}